\documentclass{article}

\usepackage{PRIMEarxiv}
\usepackage[utf8]{inputenc} 
\usepackage[T1]{fontenc}    
\usepackage{hyperref}       
\usepackage{url}            
\usepackage{booktabs}       
\usepackage{amsfonts}       
\usepackage{nicefrac}       
\usepackage{microtype}      
\usepackage{lipsum}
\usepackage{fancyhdr}       
\usepackage{graphicx}       
\usepackage{amsmath}
\usepackage{bm}
\usepackage[ruled]{algorithm2e}

\graphicspath{{media/}}     

\title{Combined Data and Deep Learning Model Uncertainties: An Application to the Measurement of Solid Fuel Regression Rate
}

\author{
  Georgios Georgalis \\
  Data Intensive Studies Center \\
  Tufts University \\
  Medford, MA 02155, USA\\
  \texttt{georgios.georgalis@tufts.edu} \\
   \And
  Kolos Retfalvi, Paul E. DesJardin \\
  Department of Mechanical and Aerospace Engineering \\
  University at Buffalo, The State University of New York, \\
  Buffalo, NY 14260, USA\\
  \texttt{\{kretfalvi, ped3\}@buffalo.edu} \\
   \And
   Abani Patra \\
   Data Intensive Studies Center and Department of Mathematics \\
   Tufts University \\
   Medford, MA 02155, USA \\
   \texttt{abani.patra@tufts.edu} \\
}

\begin{document}

\maketitle
\thispagestyle{fancy}

\begin{abstract}

In complex physical process characterization, such as the measurement of the regression rate for solid hybrid rocket fuels, where both the observation data  and the model used have uncertainties originating from multiple sources, combining these in a systematic way for  quantities of interest (QoI) remains a challenge. In this paper, we present a forward propagation uncertainty quantification (UQ) process to produce a probabilistic distribution for the observed regression rate $\dot{r}$. We characterized two input data uncertainty sources from the experiment (the distortion from the camera $U_c$ and the non-zero angle fuel placement $U_\gamma$), the prediction and model form uncertainty from the deep neural network ($U_m$), as well as the variability from the manually segmented images used for training it ($U_s$). We conducted seven case studies on combinations of these uncertainty sources with the model  form uncertainty. The main contribution of this paper is the investigation and inclusion of the experimental image data uncertainties involved, and how to include them in a workflow when the QoI is the result of multiple sequential processes.
\end{abstract}


\section{Introduction}
When analyzing a physical system, we use the ``input data'' (knowledge we have) to run the ``model'' and evaluate its outputs (knowledge we want to obtain). All of the components involved in the process carry their own assumptions, limitations, and uncertainties. When the outputs are part of a larger computational framework or are used in decision making, reporting the associated uncertainty  is required, because the modeling process and the model evaluations are approximations of the underlying true physical process. Probability models are common representations of such uncertainty in both the inputs and outputs. Forward-propagation of the uncertainty of input data and estimation of the probability distribution of the simulation output via sampling based Monte-Carlo methods or functional approximations is abundant in literature with recent applications in many fields (e.g., see \cite{MORENORODENAS201946}, \cite{TAN2021102935}, \cite{JonesUQ}). Similarly, observation data-driven calibration procedures using Markov Chain Monte Carlo or variants are common \cite{SmithUQ2014}.

However, in cases where the ``input data'' are results of a complex physical process that is inherently variable (e.g., from a complex experiment), and the ``modeling'' includes multiple sequential models and dependencies, propagating and combining all uncertainty sources in a systematic way for a required quantity of interest (QoI) remains a challenge \cite{Psaros_UQreview}. This is especially true when using operations like image segmentation models \cite{3duqcnn} for identification of  surfaces and interfaces which implicitly and explicitly use modeling and calibration. In particular, we focus here on the characterization of the ``input data" uncertainty when it is acquired in a complex experiment and used within a model-based interpretation that transforms the raw observation into a QoI. 

\begin{figure}
	\centering
  \includegraphics[width=16cm]{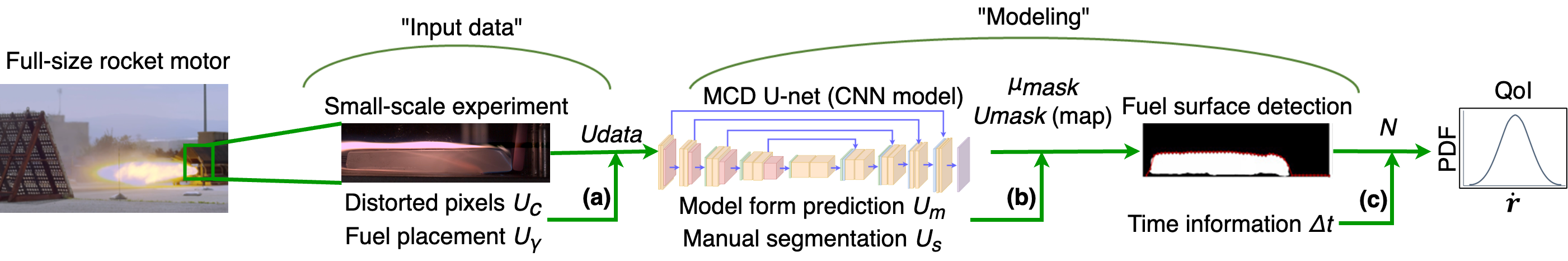}
\caption{Overview of the workflow and uncertainties at each step to estimate the fuel regression rate $\dot{r}$ from a slab burner experiment. Hybrid rocket image (left) is NASA's Peregrine rocket motor.}
\label{fig:Fig1}
\end{figure}

In hybrid rocket motor combustion, one QoI is the rate at which the fuel surface recedes during the burn, defined as the regression rate  $\dot{r}$. The regression rate has direct impact on the geometrical design of the rocket motor and its performance (\cite{Zilliac_Karabeyoglu_AIAA_RR}, \cite{Zilliac-uq}). Due to the high cost  of building a full-scale hybrid motor, it is common practice to estimate the regression rate {\it a priori} via a smaller-scale experiment of a slab burner \cite{Karabeyogluslab}. Tracking the surface of the fuel from experimental images during the burn with time data is one way to estimate $\dot{r}$ \cite{BUDZINSKI2020248}. In this case, ``inputs'' are all of the assumptions and limitations of the experiment, the equipment, and any processing to get the fuel surface images. ``Modeling'' is the process of translating the images to the regression rate, which is practically a sequence of multiple models: a deep learning convolutional network model to segment the fuel masks of an entire experiment (e.g., Monte-Carlo Dropout (MCD) U-net from \cite{SURINA2022160}), a process to detect the boundaries of the segmented fuel masks, and a process to estimate $\dot{r}$ from the changing boundaries with time data (Fig. \ref{fig:Fig1}). 

The regression rate and its measure of uncertainty are necessary to either validate multi-scale computational models (e.g., simulations of turbulent combustion chemistry coupled with flow dynamics \cite{Hawkes_2005}), or for decision making on the rocket motor design and performance. In these coupled systems, there is the additional requirement that the uncertainty information of the QoI be expressed in a simple way (e.g., as a probability distribution) because simulations can be sequential: the distribution of the regression rate represented by an ensemble is an input to another simulation (e.g., see \cite{YOUSEFIAN202123927} for an application of sequential uncertainty quantification in combustion systems).

iterature has focused on separating the different types of uncertainty within a model as aleatoric (\textit{AU}) or epistemic uncertainty (\textit{EU}) and how to measure them in physical models such as hypersonic flows \cite{Bettis_Hypersonic}, in machine learning \cite{Hullermeier_UQinML}, in Bayesian convolutional networks \cite{NIPS2017_2650d608}, and in safety decision making \cite{HORA1996217} \cite{KIUREGHIAN2009105} among others. 
In the context of a complex ``input data'' and ``modeling'' process as in the regression rate case (Fig. \ref{fig:Fig1}), these two distinct uncertainty definitions are often challenging to separate or measure. For example, the inherent camera structure and how it represents an image captured in the experimental environment and whether the researcher misplaced the fuel specimen by a few degrees include some inherent randomness, but we have the capability to estimate \textit{some} parts of that experimental randomness by analyzing the setup and the optical errors. Therefore, in this work, we do not classify uncertainty as aleatoric or epistemic, but rather characterize and quantify the individual uncertainty measures based on their source (i.e., experiment or modeling).

In this paper, we present an uncertainty quantification (UQ) process to produce a probabilistic distribution for the regression rate $\dot{r}$ from the input experimental images, by combining the input data uncertainty with inference model uncertainty. We characterize the input data uncertainties from the experiment $U_{data} = \{U_c, U_\gamma \}$, the model form uncertainty from the MCD U-net ($U_m$), and the variance in the model prediction from the variability of manually tracing the masks used to train the MCD U-net ($U_s$). The process is shown in Fig. \ref{fig:Fig1}: \textbf{(a)} We introduce the input data uncertainty $U_{data}$ to create sequenced ensembles of experimental images, \textbf{(b)} pass the ensembles through the U-net to get an estimate of the mean predicted masks $\mu_{mask}$ and corresponding uncertainty maps $U_{mask}$ which carry both the data and model uncertainties, and lastly \textbf{(c)} process each sequenced ensemble of predicted fuel masks and uncertainty maps to get an estimate for the regression rate $\dot{r}$ and its bounds.

The sources of input data uncertainty are most relevant for this problem because optical distortion from the camera ($U_c$) or non-zero angle placement of the fuel ($U_\gamma$) may misrepresent the fuel boundaries in the images that are tracked to accurately measure the regression rate. The model form uncertainty ($U_m$) is computed as the entropy of the probability prediction vector of the MCD U-net. The possible variation in the manually segmented masks used for training the MCD U-net is added as a prediction variance $U_s$ directly to the final resulting uncertainty map $U_{mask}$. 

The paper is organized as follows. Section 2 includes details about the experimental setup and the characterization for each of the studied uncertainty measures. Section 3 presents a series of cases that investigate how the different types of uncertainties impact the overall flow of uncertainty measure in the forward-propagation process (i.e., how the input data uncertainties individually and together combine with the model form uncertainty and the manual segmentation variance to form the output uncertainty map used in the measuring $\dot{r}$). Section 4 is an application of the entire workflow shown in Fig. \ref{fig:Fig1}, and includes the resulting distribution for $\dot{r}$ via forward-propagation after including all the uncertainty sources in sequence. Section 5 summarizes the conclusions of the paper with directions on future work.

\section{Formulation of Uncertainty Sources and Regression Rate Measurement}
In this section, we describe the experimental setup and equipment used, the characterization of the input data uncertainties $U_{data} = \{U_c, U_\gamma \}$ and how they are introduced to the experimental images. $U_c$ corresponds to error from optical distortion that is inherent to the camera placement in respect to the fuel specimen. $U_\gamma$ corresponds to the non-zero angle fuel placement error, which may result from the researcher not perfectly placing the fuel orthogonal to the camera axis, thus having the camera misrepresent the distance between the fuel boundaries correctly. We also show the formulation for the model form uncertainty $U_m$ and the manual segmentation variance $U_s$, as well as how the model outputs $\mu_{mask}, U_{mask}$ are used to measure the regression rate $\dot{r}$ and its bounds.

\subsection{Experimental setup and its uncertainties}

\begin{figure}
  \centering
    \includegraphics[scale = 0.5]{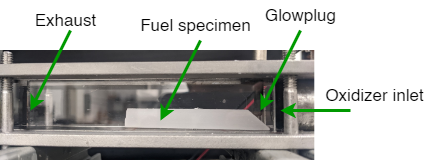}
    \caption{Slab burner test chamber with paraffin wax sample.}
\label{fig:Fig2}
\end{figure}

The experimental setup follows closely the one developed by Dunn et al. \cite{DUNN2018371} and can be seen in Fig. \ref{fig:Fig2}. The fuel specimen is placed in a chamber consisting of two stainless steel plates on the top and bottom and high temperature borosilicate glasses on the side of the experiment for optical access. The chamber is 15.24cm long, 2.54cm tall and 2.54cm wide, with a oxidizer inlet pipe of 1.83cm long and 2.54cm in diameter. Based on the entrance length, the inlet flow is assumed to be fully developed as it enters the chamber. 
We used lab-grade paraffin wax from the Carolina Biological Company as the fuel during the experiments and they were cast in a stainless steel mold. The temperature of the mold was monitored during the solidification process to avoid impurities in the samples. The average dimensions for the fuel specimens used in the experiments were 9.4mm wide, 80.7mm long and 11.2mm in height. Each sample had a 45$^{\circ}$ slant in the front to guide the flow. The oxidizer used for the experiment was 100$\% $ gaseous oxygen which was regulated with solenoids and measured with an Omega FMA 1744a mass flow meter. The measurement range of the flowmeter was 5 - 500 SLMs with $\pm 1.5 \%$ accuracy. 

\begin{figure}
  \centering
    \includegraphics[scale = 0.5]{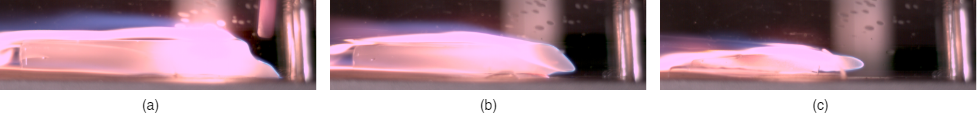}
    \caption{Example images from the experiment with oxidizer flux $G_2 = 10.96 \frac{kg}{m^s s}$ at early (a), middle (b), and later (c) stages.}
\label{fig:Fig3}
\end{figure}

To obtain the image dataset used in this paper, we conducted two experiments with measured oxidizer mass fluxes of $G_1 = 6.96 \frac{kg}{m^2 s}$ and $G_2 = 10.96 \frac{kg}{m^2 s}$ respectively.  During the automated experimental sequence the oxidizer was first introduced and the slab ignited using a Bosch diesel glow plug which was lowered to make contact with the slab and later retracted using a stepper motor. The experiment continued until the flame was extinguished. 
The experiment was recorded with a Chronos 2.1 high speed camera during the experimental runs, with a frame rate of 1000 frames per second. The lens used was a NIKON AF Nikkor 50mm lens with apertures ranging from 1.8 to 8 with an exposure of 1 $\mu s$. The full burn of the specimen took between 8-12 seconds. Some example images from the experiment with oxidizer flux $G_2 = 10.96 \frac{kg}{m^s s}$ at different times can be seen in Fig. \ref{fig:Fig3}.

\subsubsection{Optical distortion error $U_c$}
\begin{table}[!b]
\caption{Camera resolution and parameter estimation from calibration}
\centering\begin{tabular}{|c|c|}
\hline
\textbf{Camera Parameter}   & \textbf{} \\
\hline
Resolution & 1920x1080 [pix]\\
$f_x$ & 4386.76 $\pm$ 27.41 [pix] \\
$f_y$ & 4374.34 $\pm$ 26.68[pix]\\
$s$ & 0.0 (perpendicular axes) \\
$c_x$ & 971.65 $\pm$ 1.25 [pix] \\
$c_y$ & 503.64 $\pm$ 1.22 [pix] \\
$k_1$ & 4.30 $\pm$ 0.24 [--]\\
$k_2$ & 28.74 $\pm$ 13.40 [--] \\
\hline
\end{tabular}
\label{tab:cameraspecs}
\end{table}
The image data from the experiment are not exactly accurate representations of the real phenomena during the burn. One reason is that despite best efforts to set the camera correctly for best image quality, the fuel surface moves with respect to the camera during the experiment, adding a possibility of optical distortions. To account for the possible camera distortions as part of the input data uncertainty, we followed a calibration process by using checkerboard patterns. Assuming the pinhole model for the camera, i.e. the real points of the captured space are imaged by rays that pass through a single origin of the camera lens \cite{Bouguet_pinhole}, we estimated the intrinsic and extrinsic parameters of the camera and the distortion coefficients as outlined in \cite{Salazar_CameraCalib}.

Given the global coordinate system $(x,y,z)$ and the camera coordinate system $(i,j,k)$, a point in the real world is represented by the position vector $\mathbf{p} = [p_x,p_y,p_z]^T$. The same point in the camera coordinate system is represented by the position vector $\mathbf{q} = [q_i, q_j, q_k]^T = R^T(\mathbf{p}-\boldsymbol{\tau})$, where $R$ is the rotation matrix and $\boldsymbol{\tau}$ the translation vector (extrinsic parameters). The point represented in the camera coordinates as $\mathbf{q}$ is then mapped into the image plane using the intrinsic parameters matrix $K$ and represented as $\bm{r} = K \mathbf{q} = \begin{bmatrix}
f_x & 0 & 0\\
s & f_y & 0 \\
c_x & c_y & 1
\end{bmatrix} [q_i, q_j, q_k]^T$
where $f_x, f_y$ are the focal lengths, $s$ is the skew coefficient, and $c_x, c_y$ are the coordinates of the optical center. The rotation matrix $R$, the translation vector $\boldsymbol{\tau}$, and the intrinsic matrix $K$ are all estimated during the camera calibration process. In addition, we also estimate the camera distortion coefficients $k_1, k_2$ that characterize radial distortion in the image coordinates \cite{distort}: $r_{x,distorted} = \frac{r_x}{f_x}(1+k_1d^2+k_2d^4)$ and $r_{y, distorted} = \frac{r_y}{f_y}(1+k_1d^2+k_2d^4)$, where $d = (\frac{r_x}{f_x})^2+(\frac{r_y}{f_y})^2$.

\begin{figure}[!tbp]
  \centering
  \begin{minipage}[b]{0.4\textwidth}
    \includegraphics[scale = 0.3]{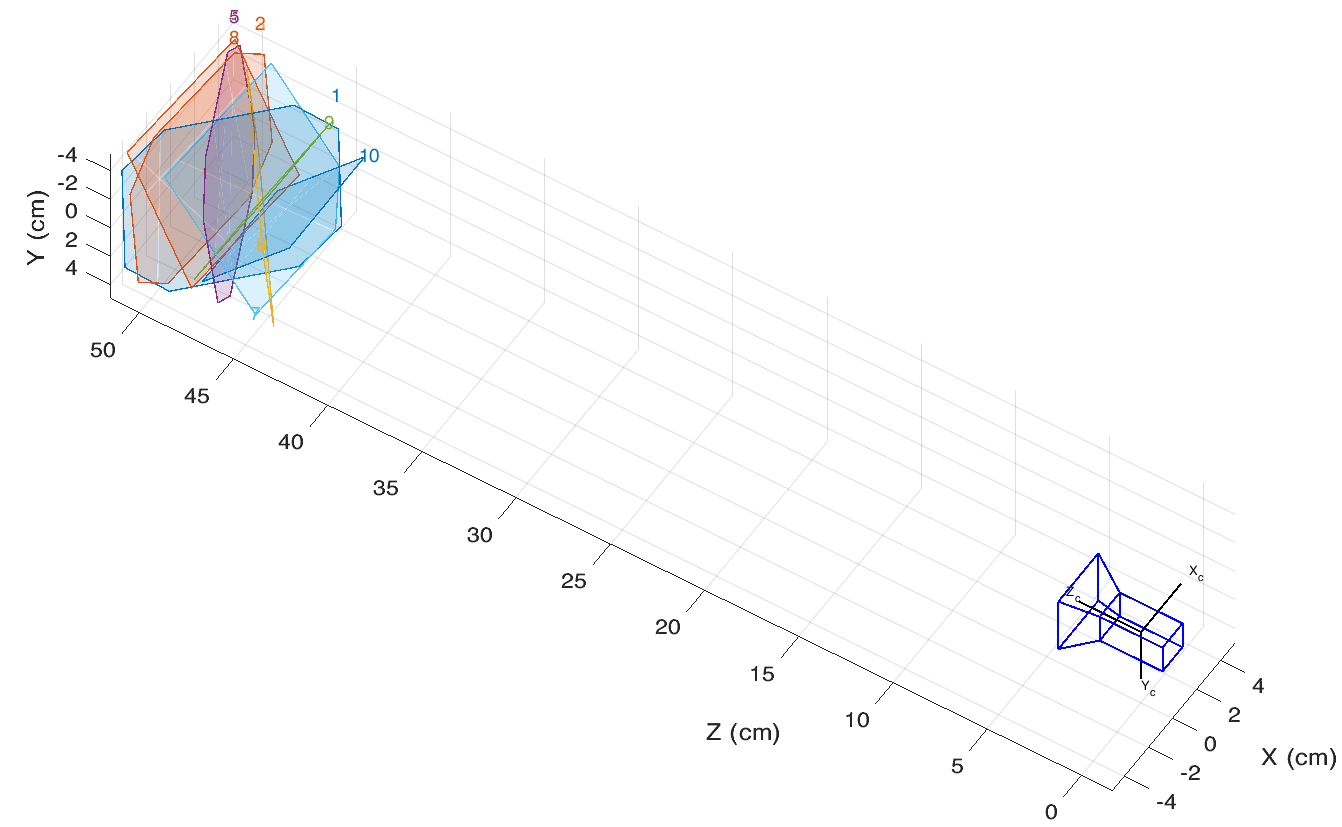}
    \caption{Relative positioning between the camera and ten 2cm-side checkboard calibration images.}
\label{fig:Fig4}
  \end{minipage}
  \hfill
    \begin{minipage}[b]{0.3\textwidth}
      \includegraphics[scale = 0.35]{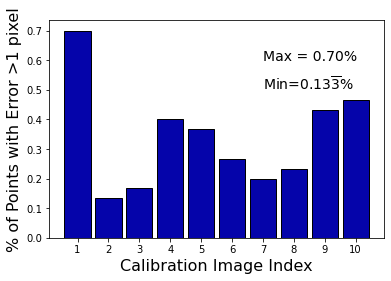}
\caption{For each of the calibration images, we expect between 0.133 to 0.7 $\%$ of the image to be distorted for a distance greater than 1 pixel. From this result, we assume a random experimental image from our camera to have a number of distorted pixels $ U_c \sim \mathcal{U}(0.133,0.7)[\%] $. }
\label{fig:Fig5}
  \end{minipage}
\end{figure}

To calibrate the camera, we took ten images of a square checkerboard pattern with a side of 2 cm under different angles (Fig. \ref{fig:Fig4}), and processed them using MATLAB's camera calibrator app \cite{bouguet2008camera}. The resulting estimated parameters and camera resolution are shown in Table \ref{tab:cameraspecs}. The rotation matrices and translation vectors are not shown because they are different for each calibration image. During the calibration process, the true locations of the checkerboard pattern are detected and compared with their locations on the reconstructed images from the camera model. The reconstruction error is then $e_{i,j} = s_{i,j} - \hat{s_{i,j}} $, where $s_{i,j}$ are the true detected points on the calibration patterns and $\hat{s_{i,j}}$ the re-constructed points. To arrive at a measure of distortion uncertainty from the camera that can be used to introduce uncertainty to the slab burner images, we count how many of the calibration points have a total error that is greater than 1 pixel in distance, for each of the calibration images. The threshold is set at 1 pixel because when we represent the experimental images as tensors for the deep learning model, an error greater than 1 pixel means the image point is distorted enough to not populate that field in the tensor anymore. The total number of distorted pixels from each calibration image are expressed as a percentage of the total pixels in the image. The result of this process is shown in Fig. \ref{fig:Fig5}: the expected maximum number of distorted points with an error greater than 1 pixel corresponds to 0.7 $\%$ of the image and the minimum number to $0.133 \%$ of the image. Therefore, based on the calibration results, we expect that an experimental image taken with our camera has a number of distorted pixels $ U_c \sim \mathcal{U}(0.133,0.7)[\%] $, expressed as a percentage of the image. Since there are only two options for the pixels, distorted or not, for a given experimental image $X_k$ of resolution 512 by 64, there is a corresponding distortion map $D_{map, k}$. The distortion map follows the binomial distribution $D_{map, k} \sim B(n = 64*512, p = U_c)$. The pixels in $X_k$ that have a success trial on the distortion map $D_{map, k}$, are distorted, and therefore have their intensity reduced to zero when we introduce distortion uncertainty $U_c$ to the original image (Alg. \ref{alg:Uc}).

\begin{algorithm}[!t]
	\caption{Add distortion uncertainty $U_c$ to experimental images}
	\label{alg:Uc}
Given experimental image $X_k: \mathbb{P}_{i,j} = \rightarrow [0,255]^3$ with $i = [0;512]$ and $ j = [0;64]$\\
	 \hspace{1em}	\textit{Step 1}. Draw a sample for the percentage of distorted pixels $ U_c \sim \mathcal{U}(0.133,0.7)[\%] $\label{step01}\\
		\hspace{1em} \textit{Step 2}.  Create the corresponding distortion map $D_{map, k} \sim B(n = 64*512, p = U_c)$  \label{step02}\\
		\hspace{1em} \textit{Step 3}.  Create the uncertain image copy $\widetilde{X_{k}} = 
            \begin{cases}
            0 & \text{for  }  D_{map, k} = 1 \\
            X_k & \text{otherwise}
            \end{cases}
  $\label{step03} \\
\end{algorithm}

\subsubsection{Non-zero angle fuel placement error $U_\gamma$}
\begin{figure}[!tbp]
  \centering
  \begin{minipage}[b]{0.4\textwidth}
    \includegraphics[scale = 0.7]{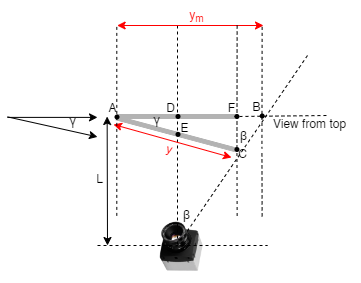}
    \caption{If the fuel specimen is not placed orthogonally to the camera, but with an offset small angle $\gamma$, the distance between the fuel boundaries is misrepresented ($y_m \ne y$).}
\label{fig:Fig6}
  \end{minipage}
  \hfill
    \begin{minipage}[b]{0.4\textwidth}
      \includegraphics[scale = 0.5]{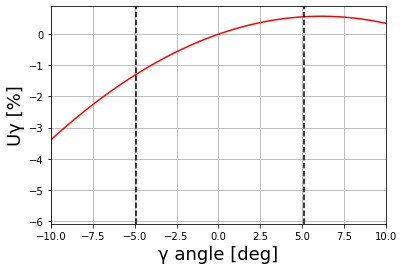}
\caption{Change of non-zero angle placement error $U_\gamma$ for different placement angles $\gamma$. For this analysis, we choose $-5^o \le \gamma \le 5^o$ to be a possible range of angle values, which corresponds to $-1.319 \le U_\gamma [\%] \le 0.558$.}
\label{fig:Fig7}
  \end{minipage}
\end{figure}

Under ideal circumstances, the slab burner is positioned perfectly orthogonal to the axis of the camera. However, the researcher that places the fuel specimen may not always do so with precision, and place the fuel at an offset small angle $\gamma \ne 0$, which then adds uncertainty to the experimental images via over- or under-estimating the distances between borders for the specimen \cite{camera_gamma}. The relative border location is important in accurate estimation of the regression rate, since we track the border changes to measure it. The schematic in Fig. \ref{fig:Fig6} shows the geometry of this phenomenon. Given the true length of the fuel specimen $y = 8.069 [cm]$, the length of the specimen to the camera $L = 37.46 [cm]$, the angle of the camera to the end of the specimen if placed correctly $\beta = tan^{-1}\frac{y}{2L} = 6.147^o$, and solving the geometry, we get an estimate for the perceived specimen length $y_m = y(cos\gamma + tan\beta sin\gamma)$. We express the error in the perceived fuel length, caused by the non-zero angle placement of the specimen, as $U_\gamma = \frac{y_m-y}{y} [\%]$. The extreme values for $U_\gamma$ are directly related to the extreme values for the angle $\gamma$. We assume that $-5^o \le \gamma \le 5^o$, because a placement by a larger angle would likely be visually noticed by the researcher and corrected before the experiment, which corresponds to $-1.319 \le U_\gamma [\%] \le 0.558$ as shown in Fig. \ref{fig:Fig7}. Given that we have no prior knowledge about the distribution of the error, we assume the non-zero angle placement uncertainty to be uniform $U_\gamma \sim \mathcal{U}(-1.319,0.558)[\%]$.

\begin{figure}
	\centering
  \includegraphics[scale = 0.5]{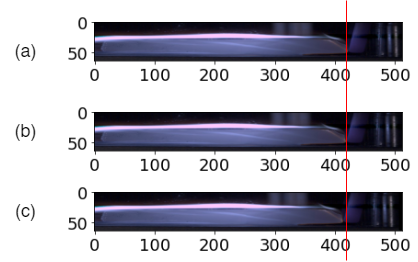}
\caption{Different versions of the same experimental image at extreme values: (a) $U_\gamma =0.558 \%$, (b) $U_\gamma = 0$, and (c) $U_\gamma = - 1.319 \%$. The red line is a reference point on the edge of the right boundary in the image with no uncertainty to aid the visualization of the difference in the cases with uncertainty present.}
\label{fig:Fig8}
\end{figure}

To introduce $U_\gamma$ to a given experimental image $X_k$, we first estimate the pixel density for the fuel specimen $\rho_k$. Given the area of the fuel surface (which can be taken from the manually segmented or predicted mask) and the length of the slab, we estimate the density $\rho_k = \frac{\text{fuel length in pixels}}{y}$ [pix/cm] for that image. We then translate $U_\gamma$ into [cm] units $U_{\gamma,cm} = U_\gamma/100 * y$, expressing the uncertainty as the length-wise error between the boundaries. Then the quantity $U_{\gamma,cm} * \rho_k$ provides the number of pixels that the right fuel boundary has to shift to introduce the sampled value of $U_\gamma$. If $U_\gamma > 0$ then the right boundary is shifted to the right, and if  $U_\gamma < 0$ then the right boundary is shifted to the left. The process of shifting includes either removing a vertical strip in the middle of the image and stitching the remaining image together adding background pixels to the right ($U_\gamma < 0$) or duplicating the vertical strip in the middle of the image and removing some of the pre-existing background on the right ($U_\gamma > 0$). The manipulation happens in the horizontal middle (at 256 pixels) since the right and left boundaries are far away from that location. The reader is referred to Alg. \ref{alg:Ug} for the process of introducing $U_\gamma$ to the images and two visual examples of the extreme cases in Fig. \ref{fig:Fig8}.

\begin{algorithm}[!t]
	\caption{Add non-zero angle placement uncertainty $U_\gamma$ to experimental images}
	\label{alg:Ug}
Given experimental image $X_k: \mathbb{P}_{i,j} = \rightarrow [0,255]^3$ with $i = [0;512]$ and $ j = [0;64]$\\
	 \hspace{1em}	\textit{Step 1}. Draw a sample for the non-zero angle uncertainty $ U_\gamma \sim \mathcal{U}(-1.319,0.558)[\%] $\label{step01}\\
		\hspace{1em} \textit{Step 2}.  Translate to length-wise units and find pixel density $U_{\gamma,cm} = U_\gamma y/100$ and $\rho_k = \frac{\text{fuel length in pixels}}{y}$
  \label{step02}\\
		\hspace{1em} \textit{Step 3}.  \textbf{If $U_\gamma > 0$}: 
            $ 
            \begin{cases}
            \text{Loc1} = \text{floor}(256-U_{\gamma,cm}\rho_k/2) \\
            \text{Loc2} = \text{floor}(256+U_{\gamma,cm}\rho_k/2) \\
            \text{Added slice} \hspace{1em} S_k = X_k[\text{loc1}:\text{loc2}] \\            
            \widetilde{X_k} = \text{append}([X_k[:256], S_k, X_k[256:512-\text{size}(S_k)]) \\
            \end{cases}
  $\label{step03} \\
  		\hspace{4em}  \textbf{If $U_\gamma < 0$}: 
            $ 
            \begin{cases}
            \text{Loc1} = \text{floor}(256-U_{\gamma,cm}\rho_k/2) \\
            \text{Loc2} = \text{floor}(256+U_{\gamma,cm}\rho_k/2) \\  
            \widetilde{X_k} = \text{append}([X_k[:\text{loc1}], X_k[\text{loc2}:512], [0]_{\text{loc2-loc1}}) \\
            \end{cases}
  $\label{step04} \\
\end{algorithm}

\subsection{Monte-Carlo Dropout (MCD) U-net for Segmentation}
\subsubsection{Model Form Uncertainty $U_m$}
\begin{figure}
	\centering
  \includegraphics[width=15cm]{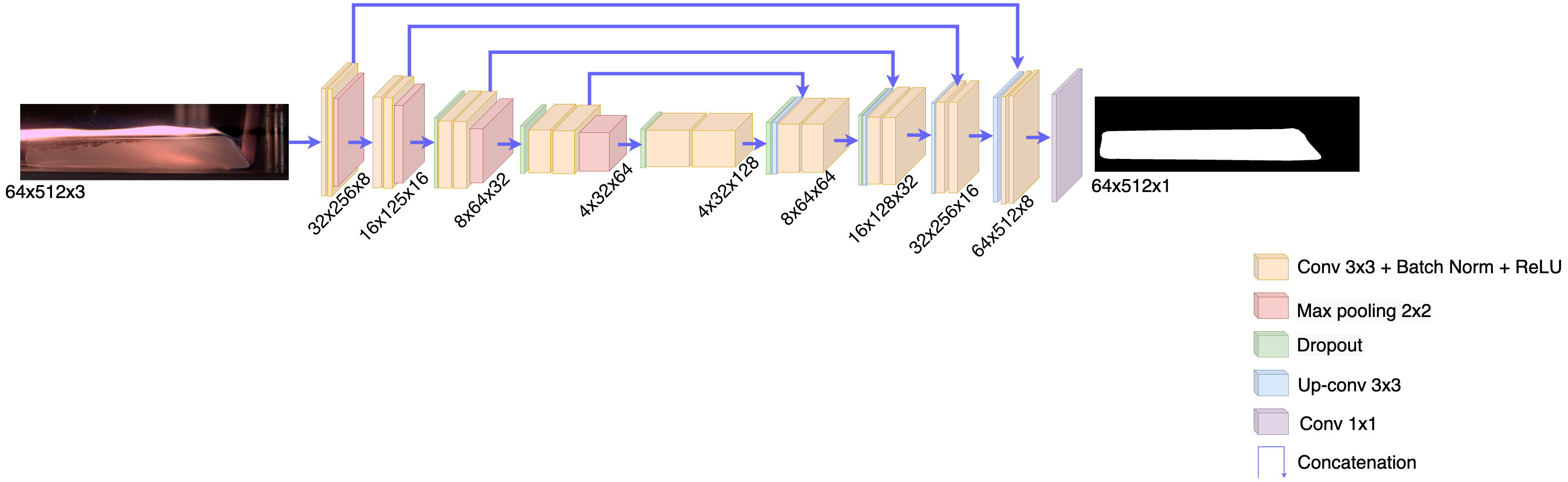}
\caption{The U-net neural network architecture with dropout in the intermediate layers. Given an input experimental image, the U-net outputs the predicted fuel segmentation mask.}
\label{fig:Fig9}
\end{figure}

We created the U-net architecture using the Keras library \cite{chollet2015keras}. Fig. \ref{fig:Fig9} shows a schematic of the U-net architecture with five dropout layers. The U-net has an encoding path that initially extracts the most important features from the image as resolution reduces and depth increases, followed by a decoding path to return to higher resolutions by reconstructing the image using up-sampling \cite{UnetOG}. The encoding path consists of four sequences of: two 3x3 convolutions, batch normalization, ReLU activation, and max pooling. The decoding path consists of four sequences of: 3x3 up-convolutions with a concatenation of the feature map from the corresponding level in the encoding path, two 3x3 convolutions, batch normalization, and lastly a ReLU activation. Finally, the U-net includes a 1x1 convolution operation to output the segmented fuel mask. The total number of trainable parameters in the network are 492,609. The parameters of a convolution layer are equal to $out \times [ in \times (3 \times 3) + 1]$, where \textit{out} is the number of features after the convolution, \textit{in} is the number of features before the convolution, the 9 parameters of the $3 \times 3$ filter, and one parameter at each node for the bias term. For example, the first convolution layer takes the original image ($64 \times 512 \times 3$), which after the convolution becomes ($64 \times 512 \times 8$), therefore that layer has $8 \times [ 3 \times (3 \times 3) + 1] =224$ parameters. Each of the batch normalization layers adds 4 additional parameters per feature of the previous layer \cite{pmlr-v37-ioffe15}. For example, since the output after the first convolution layer has 8 features, the corresponding batch normalization layer will have 32 parameters (16 of which are non-trainable). 

To estimate the prediction model form uncertainty $U_m$, we implemented Monte-Carlo Dropout (MCD), a Bayesian approximation in deep learning models \cite{mcdbayesian}. The dropout process randomly silences neurons with a probability of $p_{D}$ in the intermediate layers of the U-net. Then, at inference, the model is sampled with dropout which is an approximation to sampling from the posterior weight distribution of a fully Bayesian network \cite{ABDAR2021243}. Given a new test image $X_k$, and the training set $(X_{train}, Y_{train})$,  the resulting mean segmented mask can be approximated using Monte-Carlo integration as:
\begin{equation}
 \hat{p}(y = c|X_{k, test}, X_{train}, Y_{train})\approx \frac{1}{T}\sum_{t=1}^T \sigma(f^{\hat{W_t}}(X_k)) \label{eqn:softmax}
\end{equation}
\noindent
where $\hat{p}(y)$ corresponds to the predicted probability that a pixel in the test image $X_{k}$ is classified as $c = \{\text{Fuel}\}$, $T$ is the number samples for the weights set to 20 \cite{devries2018leveraging}, $\sigma (\cdot)$ is the softmax function, and $f$ is the model output given the weights $\hat{W_t} \sim q_\theta (W_t)$, sampled from the dropout approximating variational distribution $q_\theta (W_t)$ \cite{NIPS2017_2650d608}, which is defined for every layer $t$:
\begin{equation}
  W_t = w_t \cdot  diag([z_{t,m}]_{m = 1}^{K_t}]) 
  \end{equation}
  
  \begin{equation}
  \begin{aligned}
  z_{t,m}  \begin{cases}
            \sim Bernoulli(p_{D}) \hspace{0.5em} & \text{for layers }  t = 7, 10, 13, 15, 18 \hspace{0.5em} \text{and} \hspace{0.5em}  m = 1, ..., K_{t-1} \\
            =1 & \text{otherwise}
            \end{cases}
\end{aligned}
\end{equation}

\noindent
where $w_t$ are the weights of the trained network. The diag($\cdot$) operator maps vectors to diagonal matrices whose diagonals are the elements of the vectors \cite{derivation_dropout}. The variable $z_{t,m} = 0$ corresponds to node $m$ in layer $t-1$ to be dropped out as an input to the next layer, whereas $z_{t,m} = 1$ corresponds to the trained weights $w_t$ to be used as is, which is also the case for layers without dropout,  $p_D=0.5$ is the dropout probability, 
$K_t$ is the number of nodes/units within layer $t$. We note that the dropout probability is non-zero only in the intermediate blocks of the U-net, which corresponds to layers 7, 10, 13, 15, 18 of the network.

The model form uncertainty for the predicted probability $\hat{p}$ can be expressed as the corresponding map of the cross-entropy of the two class dimensions \cite{modelformuncertainty_book, entropy, NIPS2017_2650d608}: 
\begin{equation}
 U_m = -\sum_{c = 1}^{2}\hat{p}_{c}\log(\hat{p}_{c}) = -[\hat{p}_{_{fuel}}\log(\hat{p}_{_{fuel}})+\hat{p}_{_{background}}\log(\hat{p}_{_{background}})]\label{eqn:Um}
\end{equation}
\noindent
where $U_m$ is the model form uncertainty map, $\hat{p}_{_{fuel}}$ is the predicted probability that a given pixel corresponds to fuel, and $\hat{p}_{_{background}} $is the predicted probability that a given pixel corresponds to noise/background.

\begin{figure}
   \centering
	\includegraphics[scale = 0.5]{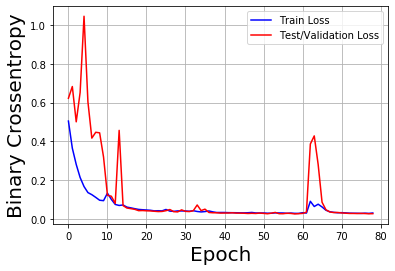}
\caption{Binary cross-entropy loss during the training of the U-net used for this work. Some abrupt changes in the loss are due to dropout effects. The training stops once the validation loss has not improved after 20 consecutive epochs.}
\label{fig:Fig10}
\end{figure}

\begin{figure}
   \centering
	\includegraphics[width = \textwidth]{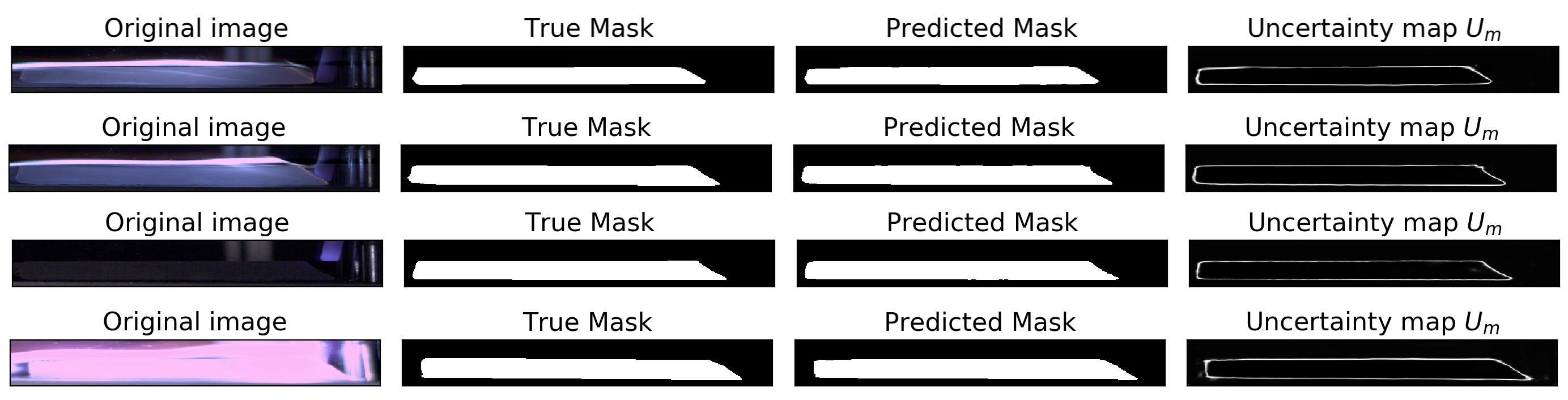}
\caption{Results from the baseline U-net model without any data uncertainty introduced to the images. The U-net performs well on all images, including low, medium, and high saturation levels during the phases of the experiment.}
\label{fig:Fig11}
\end{figure}

The U-net was trained for 79 epochs with 179 images from the early, mid, and late phases from the first experiment with oxidizer flux $G_1 = 6.96 \frac{kg}{m^2 s}$, accompanied with their manually segmented fuel masks that were traced using VGG \cite{vgg}. Fig. \ref{fig:Fig10} shows the loss progression during training. Observing the results from the testing set (Fig. \ref{fig:Fig11}), the U-net predicts the segmentation masks correctly, even for highly saturated images and without including any other components of the chamber as part of the mask (e.g., the glow plug). The uncertainty maps show the largest level of uncertainty to be around the fuel boundary geometries. The model form uncertainty further inside and outside of the identified fuel boundary is very small. 

\subsubsection{Manual segmentation uncertainty $U_s$}
In the process of training the U-net, we inherently make the assumption that the manually segmented fuel masks are ground truth. However, there is expected to be some variability between individuals that segment the fuel masks, and repeating the process multiple times to investigate the level of this variability are beyond the research scope of this paper. There have been studies to quantify the effect of improperly segmented fuel masks to the U-net outputs (e.g., see \cite{NEURIPS2020_b5d17ed2}, \cite {NEURIPS2018_473447ac}), which found that the error in the segmentation accuracy for the MNIST \cite{mnist} and MSLesion \cite{mslesion} dataset is 0.83 $\%$ and 0.56 $\%$ respectively. The MNIST dataset consists of grayscale images of digits from 0 to 9, and the MSLesion dataset of 21 medical scans. Given that the fuel segmentation masks are simple with only two labels, and without any prior knowledge, we assume the manual segmentation prediction uncertainty to be similar to those datasets and uniform  $U_s \sim \mathcal{U}(0.56,0.83) \% $, expressed as a percentage variance of the corresponding segmentation probability $\hat{p}_{_{fuel}}$. Therefore, the complete resulting uncertainty map for the mean prediction $\mu_{mask}$ of the U-net after considering all the uncertainty sources is

\begin{equation}
 U_{mask} = U_m|(U_c, U_\gamma) +\hat{p}_{_{fuel}} U_s
 \label{eqn:Umask}
\end{equation}
\noindent
where $U_{mask}$ is the complete resulting uncertainty map after all sources of uncertainty are included, $U_m $ is the model form uncertainty, $U_c$ is the distortion uncertainty, $U_\gamma $ is the non-zero angle placement uncertainty, $\hat{p}_{_{fuel}}$ is the corresponding segmentation probability for the fuel class, and $ U_s$ is the manual segmentation prediction uncertainty.

\subsection{Fuel surface detection to measure $\dot{r}$}

To approximate the regression rate from the binary fuel masks, we follow the approach described in \cite{BUDZINSKI2020248}. For a given sequence of fuel masks $\mu_{mask,k}$ representing the fuel profile between time intervals $\Delta t$, we track the height of the top surface of the fuel profile through calibration points. At each point, the localized regression rate is 
\begin{equation}
 \dot{r_{l,k}} = \frac{h(t_{k+1}) - h(t_k)}{\Delta t}
 \label{eqn:regrrate}
\end{equation}
\noindent
where $h(\cdot)$ is the fuel height for a given time measured from the bottom of the fuel mask and $\Delta t$ is the time difference between two frames. The total regression rate for the experiment is then the average localized regression rate from all points:
\begin{equation}
 \dot{r} = \frac{\sum_{k=1}^{M} \dot{r_{l,k}}}{M}
 \label{eqn:regrratetot}
\end{equation}
\noindent

Given that each of the predicted fuel masks $\mu_{mask,k}$ also has an accompanying uncertainty map $U_{mask,k}$ after the inclusion of all uncertainties, we calculate two extreme values for the total regression rate estimate $\dot{r}$. The upper bound is found by adding the uncertainty maps $U_{mask,k}$ to the corresponding fuel masks $\mu_{mask,k}$, flipping the class from ``background'' to ``fuel'' for any pixels that now have $\hat{p}_{fuel} \ge 0.5$ with the added uncertainty, and then re-calculating a total regression rate estimate $\dot{r}^+$. The lower bound comes from the opposite process: subtracting the uncertainty maps $U_{mask,k}$ from the corresponding fuel masks $\mu_{mask,k}$, flipping the class from ``fuel'' to ``background'' for any pixels that now have $\hat{p}_{fuel} < 0.5$ after the subtracted uncertainty, and re-calculating a total regression rate estimate $\dot{r}^-$. From this entire process, for a given sequence of experimental images (with or without added data uncertainties $U_c, U_\gamma$) and using the U-net (with or without the manual segmentation uncertainty $U_s$), we calculate a mean estimate for the total regression rate from the experiment $\dot{r}$ with its bounds $(\dot{r}^-,\dot{r}^+ )$.

\section{Combination and Forward-propagation of uncertainty sources}
In this section, we study all possible combinations of how each of the individual data uncertainty sources shown in Fig. \ref{fig:Fig1} impact the overall uncertainty propagation process. For each of the three uncertainty sources $U_c, U_\gamma, U_s$, we consider two scenarios: either they do not contribute (i.e., are zero) or do contribute (i.e., are uniformly distributed as described in the previous section). All possible cases are shown in the Table \ref{tab:scenarios}.  

\begin{table}[!b]
\caption{Studied cases of uncertainty sources}
\centering\begin{tabular}{|c|c|c|c|}
\hline
\textbf{Case $\#$}   & Distortion \textbf{$U_c$} & Non-zero angle \textbf{$U_\gamma$} & Manual segmentation \textbf{$U_s$} \\
\hline
Baseline & 0&0&0\\
1 & $\mathcal{U}(0.133 , 0.7) \% $ & 0 &0 \\
2 & 0  & $\mathcal{U}(-1.319 , 0.558) \% $ &0 \\
3 & $\mathcal{U}(0.133 , 0.7) \% $ & $\mathcal{U}(-1.319 , 0.558) \% $ &0 \\
4 & 0  & 0 & $\mathcal{U}(0.56 , 0.83) \% $ \\
5 & $\mathcal{U}(0.133 , 0.7) \% $ & 0 &$\mathcal{U}(0.56 , 0.83) \% $ \\
6 & 0 & $\mathcal{U}(-1.319 , 0.558) \% $ &$\mathcal{U}(0.56 , 0.83) \% $ \\
7 & $\mathcal{U}(0.133 , 0.7) \% $ & $\mathcal{U}(-1.319 , 0.558) \% $ &$\mathcal{U}(0.56 , 0.83) \% $ \\
\hline
\end{tabular}
\label{tab:scenarios}
\end{table}

\begin{figure}
  \centering
    \includegraphics[width = \textwidth]{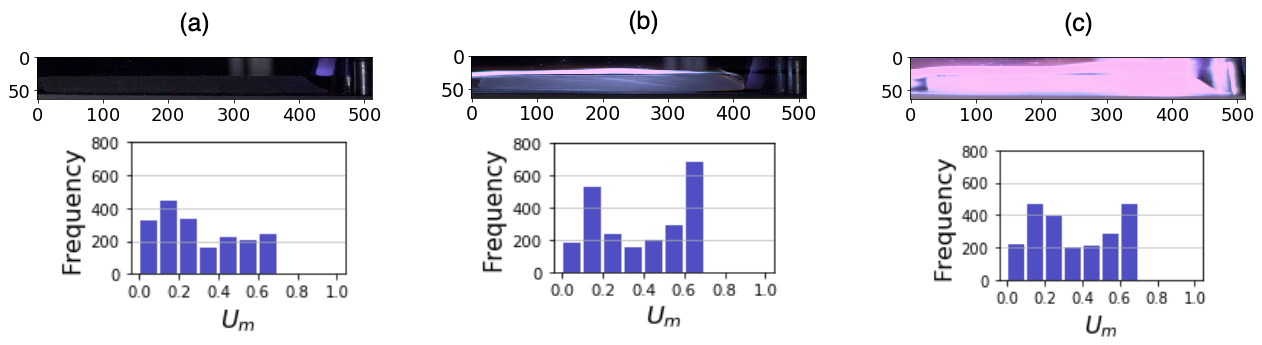}
    \caption{Example images from the experiment with different saturation levels at low (a), medium (b), and high (c) with their corresponding model form uncertainty histograms. We used these representative images to account for different image qualities, since the uncertainty distribution has a different profile for each of them.}
\label{fig:Fig12}
\end{figure}

First, we observe how the error from the camera distortion $U_c$ and the non-zero angle placement error $U_\gamma$ impact the U-net model form uncertainty map $U_m$ separately and together (cases 1,2, and 3). Secondly, we observe how the manual segmentation error $U_s$ changes the complete uncertainty map $U_{mask}$ individually and together with $U_c, U_\gamma$ (cases 4, 5, and 6). Lastly, we showcase an example where all three uncertainties are included (case 7), which is representative of the approach followed in the calculation of the full probability distribution of the regression rate in the next section, which is the end goal of the paper. For these studies, we used three representative experimental images from the testing set of the U-net that correspond to three different image saturation levels in the experiment: ''low``, ''mid``, and ''high``, as shown in Fig. \ref{fig:Fig12}. The histograms are the model form uncertainty near the boundary for each image. Observing the model form uncertainty \textit{near} the fuel boundary, we identify three different modes depending on the saturation level in the experimental images. Images with low saturation ( (a) in Fig. \ref{fig:Fig12}), show an approximately uniform profile across all uncertainty ranges. Images of medium saturation ( (b) in Fig. \ref{fig:Fig12}) show a bi-modal behavior for $U_{m}$ with a uniform profile in between the peaks. Lastly, images with high saturation ( (c) in in Fig. \ref{fig:Fig12}) has again a bi-modal profile similar to the medium saturation images, but the intermediate levels of uncertainty are higher, which is attributed to the higher saturation from the more violent burn.

We follow a standard forward-propagation uncertainty approach: If $U_c$ and $U_\gamma$ are present, we create an ensemble of perturbed/uncertain copies of a given image by following the process in Alg. 1 or Alg. 2 respectively, and pass these uncertain copies through the U-net to get the model form uncertainty map $U_m$. If $U_s$ is present, that is added to the model form uncertainty map $U_m$ proportionally to the predicted probability of each pixel (Eqn. \ref{eqn:Umask}). For each of these cases, we sampled the corresponding uniform distributions $N = 10,000$ times and show the resulting uncertainty distribution compared to the baseline case where no uncertainties are present. 

\begin{figure}
  \centering
    \includegraphics[width = 16cm]{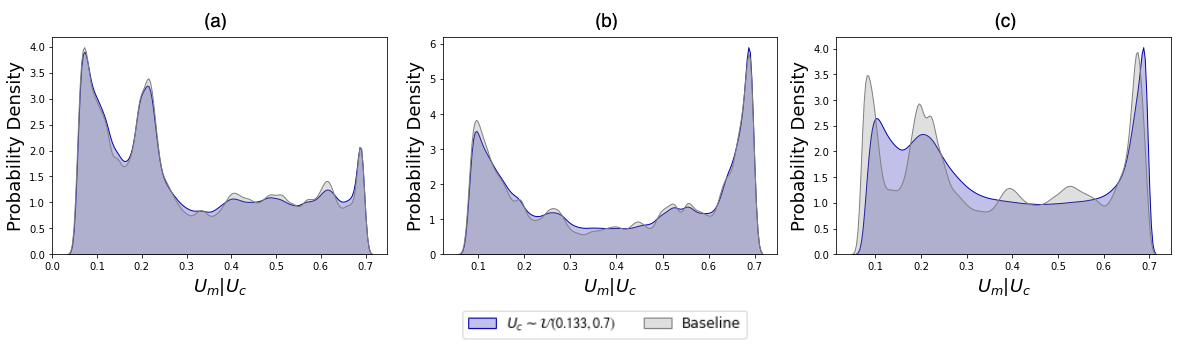}
    \includegraphics[scale = 0.5]{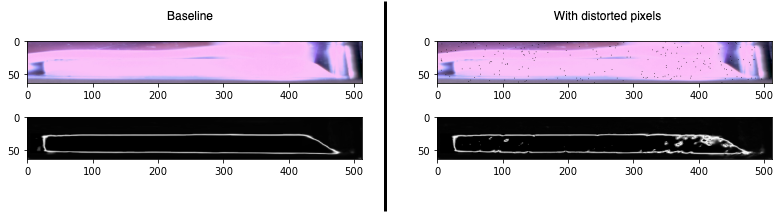}
    \caption{Case 1: Impact of the distorted pixel uncertainty $U_c$ to the model form uncertainty $U_m$ for three levels of saturation for the experimental images: low (a), medium (b), and high (c). For the image with high saturation, distortion changes the uncertainty profile for the lower uncertainty values. }
\label{fig:Fig13}
\end{figure}

For case 1, where only the uncertainty from the distorted pixels $U_c$ is present, we find that distortion adds some variability to the $U_m$ probability distribution, but still follows the overall pattern of the distribution without distortion in the low and medium saturated images. The small variability is expected since the distortion is added uniformly to the entire image. However, in the highly saturated case, distortion appears to be a moderating factor to the large peaks of small uncertainty in the original image. The explanation for this irregularity is justified in Fig. \ref{fig:Fig13}. When distortion is present in the highly saturated experimental image, the uncertainty around distorted points with high density is much larger compared to the image without distortion. The distorted points outside the fuel boundary do not contribute to the uncertainty since they were already part of the background. However, the distorted pixels that are present within the fuel boundary, add additional uncertainty, which changes the bimodal profile for the probability density present in the baseline case to a more uniform profile for the lower uncertainty values.

\begin{figure}
  \centering
    \includegraphics[width = 16cm]{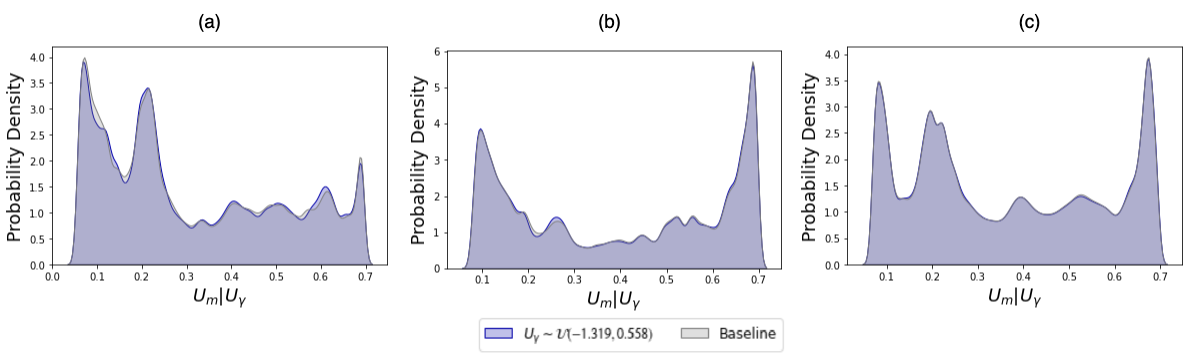}
    \caption{Case 2: Impact of the non-zero angle fuel placement uncertainty $U_\gamma$ to the model form uncertainty $U_m$ for three levels of saturation for the experimental images: low (a), medium (b), and high (c). The overall distribution of uncertainty is maintained.}
\label{fig:Fig14}
\end{figure}

For case 2, where only the non-zero angle fuel placement uncertainty $U_\gamma$ is present, we find that some variability is added, but the overall pattern of the distribution of uncertainty compared to the baseline case is maintained for all saturation levels (Fig. \ref{fig:Fig14}). Case 3, which includes both data uncertainties $U_c, U_\gamma$, also appears to maintain the overall profile, but the non-zero angle effects appear to negate some of the variability caused by distortion, likely due to the removal or addition of parts of the same image. 

\begin{figure}
  \centering
    \includegraphics[width = 16cm]{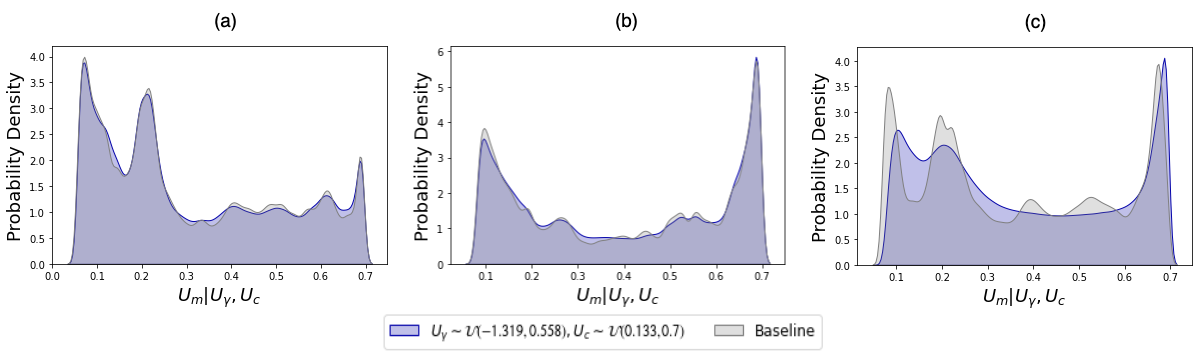}
    \caption{Case 3: Impact of the non-zero angle fuel placement uncertainty $U_\gamma$ and distorted pixel uncertainty $U_c$ to the model form uncertainty $U_m$ for three levels of saturation for the experimental images: low (a), medium (b), and high (c). Compared to case 1, the presence of non-zero angle uncertainty $U_\gamma$ negates some of the variability caused by the distortion uncertainty $U_c$.}
\label{fig:Fig15}
\end{figure}

\begin{figure}
  \centering
    \includegraphics[width = 16cm]{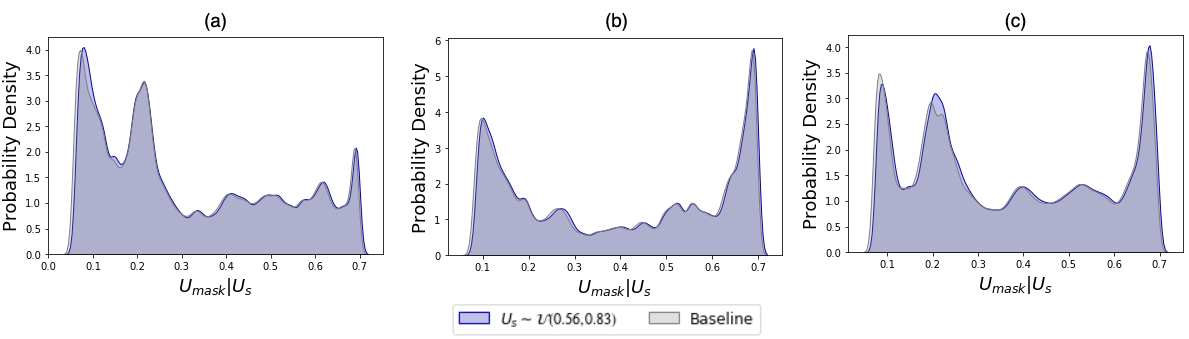}
    \caption{Case 4: Impact of the manual segmentation uncertainty $U_s$ to the overall uncertainty map $U_{mask}$ for three levels of saturation for the experimental images: low (a), medium (b), and high (c). The added effect of this uncertainty appears as a right-shift to the uncertainty distribution profile.}
\label{fig:Fig16}
\end{figure}
\begin{figure}
  \centering
    \includegraphics[width = 16cm]{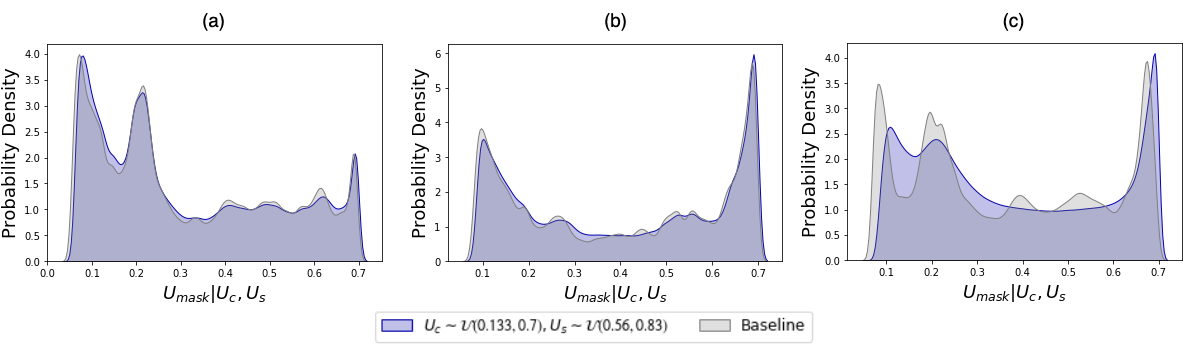}
    \caption{Case 5: Impact of the distorted pixel uncertainty $U_c$ and the manual segmentation uncertainty $U_s$ to the overall uncertainty map $U_{mask}$ for three levels of saturation for the experimental images: low (a), medium (b), and high (c).}
\label{fig:Fig17}
\end{figure}
\begin{figure}
  \centering
    \includegraphics[width = 16cm]{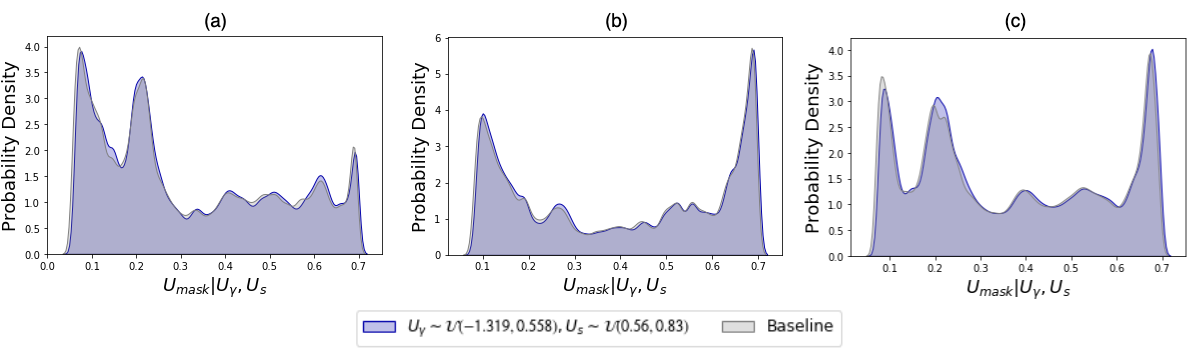}
    \caption{Case 6: Impact of the non-zero angle fuel placement uncertainty $U_\gamma$ and the manual segmentation uncertainty $U_s$ to the overall uncertainty map $U_{mask}$ for three levels of saturation for the experimental images: low (a), medium (b), and high (c).}
\label{fig:Fig18}
\end{figure}

Cases 4--6 show the effects of the manual segmentation uncertainty $U_s$ individually and together with the other uncertainty sources. $U_s$ has an added effect based on the predicted probability of a given pixel, therefore the additive effects are present for those pixels with high fuel probability (i.e., within and on the fuel boundary). The result appears as a right-shift to the uncertainty distribution profile (Figs. \ref{fig:Fig16}, \ref{fig:Fig17}, \ref{fig:Fig18}).

\begin{figure}
  \centering
    \includegraphics[width = 16cm]{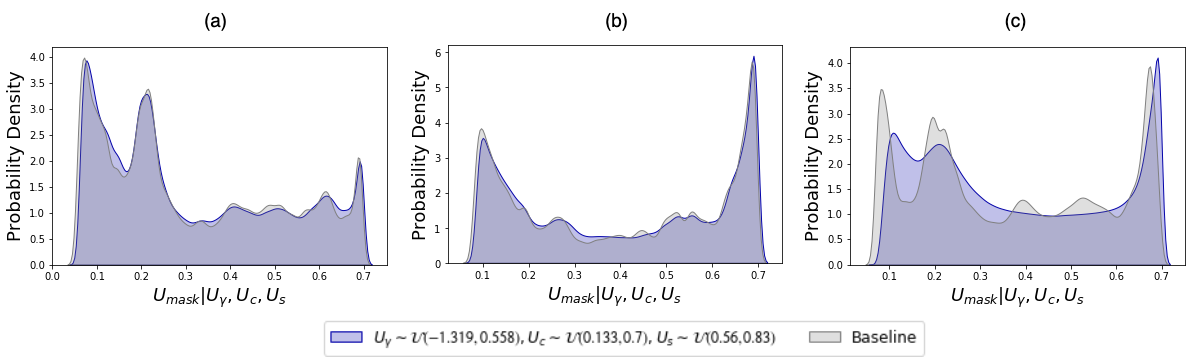}
    \caption{Case 7: Impact of the distorted pixel uncertainty $U_c$, the non-zero angle fuel placement uncertainty $U_\gamma$, and the manual segmentation uncertainty $U_s$ to the overall uncertainty map $U_{mask}$ for three levels of saturation for the experimental images: low (a), medium (b), and high (c).}
\label{fig:Fig19}
\end{figure}

Case 7 shows the combined effects of the studied uncertainties, that appear to be mixed: a right-shift caused by the manual segmentation uncertainty $U_s$, a moderating effect caused by removing or replicating parts of these images from $U_\gamma$, and added variability (and profile change for the high-saturated images) due to more areas with uncertainty from $U_c$ (Fig. \ref{fig:Fig19}). 

\section{Estimation of the distribution of $\dot{{r}}$}
\begin{figure}
  \centering
    \includegraphics[width = 16cm]{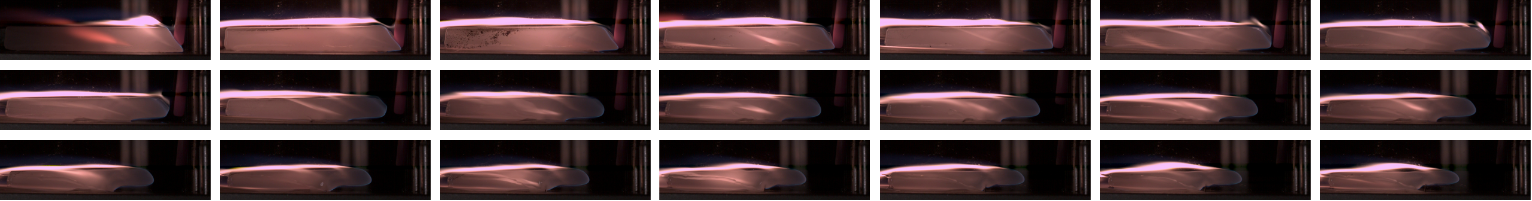}
    \caption{Images from experiment 2 with with oxidizer mass flux $G_2 = 10.96 \frac{kg}{m^2 s}$. The frames shown are taken at a fixed interval of 0.32 seconds. The sequence of the burn starts from the top left and ends at the bottom right. The goal is to measure the probability distribution of the average regression rate, the quantity that describes how fast the fuel recedes during the burn.}
\label{fig:Fig20}
\end{figure}

The goal of this section is to implement the framework shown in Fig. \ref{fig:Fig1} for the experiment with oxidizer mass flux $G_2 = 10.96 \frac{kg}{m^2 s}$, and find the probability distribution of the QoI: the total regression rate $\dot{r}$. To begin, we first capture a sequence of images from the experiment at fixed frame intervals (Fig. \ref{fig:Fig20}).

\begin{figure}
  \centering
    \includegraphics[scale = 0.4]{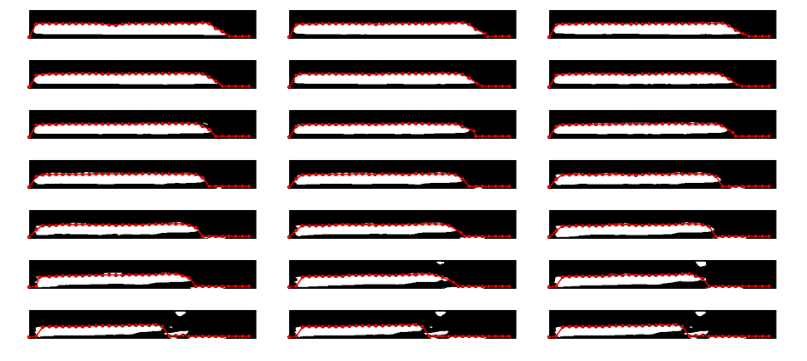}
    \caption{One ensemble of predicted fuel masks from experiment 2 with with oxidizer mass flux $G_2 = 10.96 \frac{kg}{m^2 s}$, with visible fuel tracing. The process of calibration points ensures that in cases where the U-net misclassifies some part of the picture as fuel, which is likely in the later parts of the experiment, the algorithm approximates the boundary correctly for the regression rate calculation.}
\label{fig:Fig21}
\end{figure}

Then, we follow a forward-propagation Monte-Carlo approach: the data uncertainties $U_c, U_\gamma$ are uniformly sampled and added to the sequence of images from experiment 2 (Fig. \ref{fig:Fig20}), the images are passed through the U-net to get the segmented fuel masks $\mu_{mask,k}$ , the manual segmentation uncertainty $U_s$ is uniformly sampled and added to the model form uncertainty $U_m$ (Eqn. \ref{eqn:Umask}) to get the overall uncertainty maps $U_{mask,k}$ for all the experimental images (see Fig. \ref{fig:Fig21} for one example of an ensemble of fuel masks with profile tracking). From one sequenced ensemble of  $\mu_{mask,k}$ and $U_{mask,k}$, one point measurement of the total regression rate $\dot{r}$ is computed together with its bounds ($\dot{r}^+, \dot{r}^-)$. The process is repeated until the variance of the final probability distribution of the regression rate $\sigma_{\dot{r}}^2$ converges to a tolerance of $e^{-6}$ (Alg. \ref{alg:FinalR}). 

\begin{figure}
  \centering
    \includegraphics[scale = 0.5]{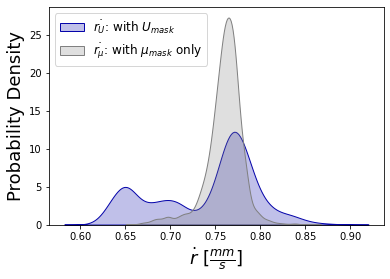}
    \caption{The probability distribution of the regression rate for the experiment with oxidizer mass flux $G_2 = 10.96 \frac{kg}{m^2 s}$. The grey curve only considers the mean fuel masks $\mu_{mask} $ for the calculation (i.e., does not include the regression rate bounds using $U_{mask}$), whereas the blue curve includes the information from the uncertainty maps. The mean of the probability distribution from the mean masks only is $\mu_{\dot{r_\mu}} = 0.759954 \frac{mm}{s}$ (grey curve) and considering the uncertainty maps as well is $\mu_{\dot{r_U}} = 0.740032 \frac{mm}{s}$. The uncertainty maps translate to signficant variability to both tail-ends of the distribution.}
\label{fig:Fig22}
\end{figure}

The distribution of the regression rate is shown in Fig. \ref{fig:Fig22}. We compare two instances: a probability distribution if we consider the uncertainty maps $U_{mask}$ that provide the bounds for the regression rate with each ensemble or if we do not. The probability distribution considering only the mean masks $\mu_{mask}$ from the ensembles resembles a Gaussian distribution with a mean regression rate $\mu_{\dot{r_\mu}} = 0.759954 \frac{mm}{s}$. Considering the uncertainty maps that provide the bounds for the regression rate from each ensemble, adds variability to both tail-ends of the distribution with a shift of the mean to $\mu_{\dot{r_U}} = 0.740032 \frac{mm}{s}$. In previous work where we only considered the model form uncertainty $U_m$, we found the regression rate for an experiment with $G = 9.58 \frac{kg}{m^2 s}$ to be $0.74 \pm 0.09 \frac{mm}{s}$. We note that when the uncertainty maps are included in the calculation, a significant amount of density is shifted towards the smaller regression rate values, because of how the surface tracking algorithm works. The surface tracking algorithm accounts for a physical constraint: it does not allow for the fuel surface to increase in size during the burn, which would be a violation of combustion laws. Some images in the later stages of the experiment ensembles can often result in larger fuel surface than previous time steps when we add the uncertainty masks, and that effect is neglected. Uncertainty masks tend to have higher values as the burn proceeds (images are more saturated). In the subtracting case, the equivalent scenario does not happen; the higher uncertainty maps in the later stages of the burn will be considered normally. Therefore, the process penalizes the higher regression rates when uncertainties are added, resulting in a more significant density shift to lower values. The contributions of this paper agree with trends from previous literature, but add the additional step of characterization and quantification of data uncertainty which translates to more accurate representation of the distribution of the regression rate when measured from images.

\begin{algorithm}[!t]
	\caption{Estimation of the distribution of the regression rate $\dot{r}$}
	\label{alg:FinalR}
Given a sequence of experimental images $X_k: \mathbb{P}_{i,j} = \rightarrow [0,255]^3$ with $i = [0;512]$ and $ j = [0;64]$\\
	 \hspace{1em}	\textbf{While} $\sigma_{\dot{r}}^{2, new} - \sigma_{\dot{r}}^{2, old} > e^{-6}$ \textbf{do}: \\
  \hspace{2em} \textit{Step 1}. Draw a sample for the non-zero angle uncertainty $ U_\gamma \sim \mathcal{U}(-1.319,0.558)[\%] $\label{step01}\\
		\hspace{2em} \textit{Step 2}.  Draw a sample for the distortion uncertainty $ U_c \sim \mathcal{U}(0.133,0.7)[\%] $  \label{step02}\\
		\hspace{2em} \textit{Step 3}.  Introduce $U_c, U_\gamma$ to the ensemble $X_k$ (Alg. \ref{alg:Uc} and Alg. \ref{alg:Ug}) \label{step03} \\
  		\hspace{2em} \textit{Step 4}. Pass the ensemble through U-net to get the mean predicted fuel masks $\mu_{mask,k}$ and the mean model form uncertainty maps $U_{m,k}$. \label{step04} \\
      \hspace{2em} \textit{Step 5}. Draw a sample for the manual segmentation uncertainty $ U_s \sim \mathcal{U}(0.56,0.83)[\%] $\label{step05}\\
            \hspace{2em} \textit{Step 6}. Compute overall uncertainty map for the ensemble $U_{mask,k} = U_{m,k}|(U_c, U_\gamma) +\hat{p}_{_{fuel,k}} U_s. $\label{step06}\\
            \hspace{2em} \textit{Step 7}. Compute point estimate for regression rate $\dot{r}$ and bounds $\dot{r}^+, \dot{r}^-$ from $\mu_{mask,k}$ and $U_{mask,k}$  \label{step07}\\
        \hspace{2em} \textit{Step 8}. Append probability distribution $p(\dot{r}) =\mathrel{+} \{\dot{r} , \dot{r}^+, \dot{r}^- \}$  and update variance $\sigma_{\dot{r}}^{2,new} $ \label{step08}\\
        \hspace{1em}	\textbf{End While} \\            
\end{algorithm}

\section{Conclusions and Future Work}
In this work, we presented a full forward-propagation framework for uncertainty quantification of the regression rate from experimental images. We characterized uncertainty sources that are pertinent to this case: distortion uncertainty $U_c$ from the camera, non-zero angle placement of the fuel $U_\gamma$, manual segmentation training uncertainty $U_s$, and model form uncertainty from the Monte-Carlo Dropout U-net $U_m$. The image data generated for this work came from two slab burner experiments with measured oxidizer mass fluxes of $G_1 = 6.96 \frac{kg}{m^2 s}$ and $G_2 = 10.96 \frac{kg}{m^2 s}$. We conducted seven case studies of all possible combinations between the three uncertainty sources and found that their effects combine: a right-shift in the model form uncertainty distribution caused by the additive effect of the manual segmentation uncertainty $U_s$, a moderating effect from $U_\gamma$ caused by the effect of placing the fuel under a small angle, and the overall added variability throughout the model form uncertainty distribution (and profile change for the high-saturated images) due to more areas with uncertainty caused from the distortion uncertainty $U_c$. We completed our study by using the images from the experiment with oxidizer flux $G_2 = 10.96 \frac{kg}{m^2 s}$, generating a large ensemble of their uncertain copies by sampling the three uncertainty sources, and tracking the fuel profiles in each case to estimate the full regression rate probability distribution. The probability distribution considering only the mean masks $\mu_{mask}$ from the ensembles resembles a Gaussian distribution with a mean regression rate $\mu_{\dot{r_\mu}} = 0.759954 \frac{mm}{s}$. Considering the uncertainty maps that provide the bounds for the regression rate from each ensemble, adds variability to both tail-ends of the distribution with a shift of the mean to $\mu_{\dot{r_U}} = 0.740032 \frac{mm}{s}$.

In future work, we will use the resulting probability distribution of the regression rate from experimental data shown in this work, together with experiments on other oxidizer fluxes, to validate an equivalent estimate for the regression rate from coupled flow and chemistry simulations for the slab burner combustion phenomena using our open-source software framework ABLATE (\url{https://ablate.dev/}). 

\section*{Acknowledgments}
The authors would like to thank the reviewer comments that greatly improved the manuscript. 

This work was supported by the United States Department of Energy’s (DoE) National Nuclear Security Administration (NNSA) under the Predictive Science Academic Alliance Program III (PSAAP III) at the University at Buffalo, under contract number DE-NA0003961.

Computational resources: The authors acknowledge the Tufts University High Performance Compute Cluster which was used for training the U-net models and running the simulations for the uncertainty analysis.

\bibliographystyle{unsrt}  
\bibliography{references}

\begin{thebibliography}{10}

\bibitem{MORENORODENAS201946}
Antonio~M. Moreno-Rodenas, Franz Tscheikner-Gratl, Jeroen~G. Langeveld, and
  Francois~H.L.R. Clemens.
\newblock Uncertainty analysis in a large-scale water quality integrated
  catchment modelling study.
\newblock {\em Water Research}, 158:46--60, 2019.

\bibitem{TAN2021102935}
Jingye Tan, Umberto Villa, Nima Shamsaei, Shuai Shao, Hussein~M. Zbib, and
  Danial Faghihi.
\newblock A predictive discrete-continuum multiscale model of plasticity with
  quantified uncertainty.
\newblock {\em International Journal of Plasticity}, 138:102935, 2021.

\bibitem{JonesUQ}
Reese~E. Jones, Michael~T. Redle, Hemanth Kolla, and Julia~A. Plews.
\newblock A minimally invasive, efficient method for propagation of full-field
  uncertainty in solid dynamics.
\newblock {\em International Journal for Numerical Methods in Engineering},
  122(23):6955--6983, 2021.

\bibitem{SmithUQ2014}
Ralph Smith.
\newblock {\em Uncertainty Quantification}.
\newblock SIAM, 2014.

\bibitem{Psaros_UQreview}
Apostolos~F Psaros, Xuhui Meng, Zongren Zou, Ling Guo, and George~Em
  Karniadakis.
\newblock Uncertainty quantification in scientific machine learning: Methods,
  metrics, and comparisons, 2022.

\bibitem{3duqcnn}
Michael~C. Krygier, Tyler LaBonte, Carianne Martinez, Chance Norris, Krish
  Sharma, Lincoln~N. Collins, Partha~P. Mukherjee, and Scott~A. Roberts.
\newblock Quantifying the unknown impact of segmentation uncertainty on
  image-based simulations.
\newblock {\em Nature Communications}, 12:5414, 2021.

\bibitem{Zilliac_Karabeyoglu_AIAA_RR}
G.~Zilliac and M.~Karabeyoglu.
\newblock Hybrid rocket fuel regression rate data and modeling.
\newblock {\em In 42nd AIAA/ASME/SAE/ASEE Joint Propulsion Conference and
  Exhibit}, 2006.

\bibitem{Zilliac-uq}
Greg Zilliac, George~T. Story, Ashley~C. Karp, Elizabeth~T. Jens, and George
  Whittinghill.
\newblock Combustion efficiency in single port hybrid rocket engines.
\newblock {\em In AIAA Propulsion and Energy 2020 Forum}, 2020.

\bibitem{Karabeyogluslab}
M.~Karabeyoglu, B.~Cantwell, and D.~Altman.
\newblock Development and testing of paraffin-based hybrid rocket fuels.
\newblock {\em 37th Joint Propulsion Conference and Exhibit}, 2001.

\bibitem{BUDZINSKI2020248}
Kenneth Budzinski, Siddhant~S. Aphale, Elektra~Katz Ismael, Gabriel Surina, and
  Paul~E. DesJardin.
\newblock Radiation heat transfer in ablating boundary layer combustion theory
  used for hybrid rocket motor analysis.
\newblock {\em Combustion and Flame}, 217:248--261, 2020.

\bibitem{SURINA2022160}
Gabriel Surina, Georgios Georgalis, Siddhant~S. Aphale, Abani Patra, and
  Paul~E. DesJardin.
\newblock Measurement of hybrid rocket solid fuel regression rate for a slab
  burner using deep learning.
\newblock {\em Acta Astronautica}, 190:160--175, 2022.

\bibitem{Hawkes_2005}
Evatt~R Hawkes, Ramanan Sankaran, James~C Sutherland, and Jacqueline~H Chen.
\newblock Direct numerical simulation of turbulent combustion: fundamental
  insights towards predictive models.
\newblock {\em Journal of Physics: Conference Series}, 16:65--79, jan 2005.

\bibitem{YOUSEFIAN202123927}
Sajjad Yousefian, Gilles Bourque, and Rory~F.D. Monaghan.
\newblock Bayesian inference and uncertainty quantification for
  hydrogen-enriched and lean-premixed combustion systems.
\newblock {\em International Journal of Hydrogen Energy}, 46(46):23927--23942,
  2021.

\bibitem{Bettis_Hypersonic}
Benjamin Bettis and Serhat Hosder.
\newblock Uncertainty quantification in hypersonic reentry flows due to
  aleatory and epistemic uncertainties.
\newblock {\em In 49th AIAA Aerospace Sciences Meeting including the New
  Horizons Forum and Aerospace Exposition}, 2011.

\bibitem{Hullermeier_UQinML}
Eyke H\"{u}llermeier and Willem Waegeman.
\newblock Aleatoric and epistemic uncertainty in machine learning: an
  introduction to concepts and methods.
\newblock {\em Machine Learning}, 110(3):457–506, 2021.

\bibitem{NIPS2017_2650d608}
Alex Kendall and Yarin Gal.
\newblock What uncertainties do we need in bayesian deep learning for computer
  vision?
\newblock In {\em Advances in Neural Information Processing Systems},
  volume~30, 2017.

\bibitem{HORA1996217}
Stephen~C. Hora.
\newblock Aleatory and epistemic uncertainty in probability elicitation with an
  example from hazardous waste management.
\newblock {\em Reliability Engineering and System Safety}, 54(2):217--223,
  1996.

\bibitem{KIUREGHIAN2009105}
Armen~Der Kiureghian and Ove Ditlevsen.
\newblock Aleatory or epistemic? does it matter?
\newblock {\em Structural Safety}, 31(2):105--112, 2009.

\bibitem{DUNN2018371}
C.~Dunn, G.~Gustafson, J.~Edwards, T.~Dunbrack, and C.~Johansen.
\newblock Spatially and temporally resolved regression rate measurements for
  the combustion of paraffin wax for hybrid rocket motor applications.
\newblock {\em Aerospace Science and Technology}, 72:371--379, 2018.

\bibitem{Bouguet_pinhole}
J.Y. Bouguet and P.~Perona.
\newblock Camera calibration from points and lines in dual-space geometry.
\newblock {\em In Proceedings of the 5th European Conference on Computer
  Vision}, pages 2--6, Jun 1998.

\bibitem{Salazar_CameraCalib}
R.~Juarez-Salazar, J.~Zheng, and Victor~H. Diaz-Ramirez.
\newblock Distorted pinhole camera modeling and calibration.
\newblock {\em Applied Optics}, 59(36):11310--11318, 2020.

\bibitem{distort}
J.~Heikkila and O.~Silven.
\newblock A four-step camera calibration procedure with implicit image
  correction.
\newblock In {\em Proceedings of IEEE Computer Society Conference on Computer
  Vision and Pattern Recognition}, pages 1106--1112, 1997.

\bibitem{bouguet2008camera}
Jean-Yves Bouguet.
\newblock Camera calibration toolbox for matlab (2008).
\newblock {\em URL http://www. vision. caltech. edu/bouguetj/calib\_doc}, 1080,
  2008.

\bibitem{camera_gamma}
{Robbe, C.}, {Nsiampa, N.}, {Oukara, A.}, and {Papy, A.}
\newblock Quantification of the uncertainties of high-speed camera
  measurements.
\newblock {\em Int. J. Metrol. Qual. Eng.}, 5(2):201, 2014.

\bibitem{chollet2015keras}
Fran\c{c}ois Chollet et~al.
\newblock Keras.
\newblock \url{https://keras.io}, 2015.

\bibitem{UnetOG}
Olaf Ronneberger, Philipp Fischer, and Thomas Brox.
\newblock U-net: Convolutional networks for biomedical image segmentation.
\newblock In Nassir Navab, Joachim Hornegger, William~M. Wells, and
  Alejandro~F. Frangi, editors, {\em Medical Image Computing and
  Computer-Assisted Intervention -- MICCAI 2015}, pages 234--241, Cham, 2015.
  Springer International Publishing.

\bibitem{pmlr-v37-ioffe15}
Sergey Ioffe and Christian Szegedy.
\newblock Batch normalization: Accelerating deep network training by reducing
  internal covariate shift.
\newblock In Francis Bach and David Blei, editors, {\em Proceedings of the 32nd
  International Conference on Machine Learning}, volume~37 of {\em Proceedings
  of Machine Learning Research}, pages 448--456, Lille, France, 07--09 Jul
  2015. PMLR.

\bibitem{mcdbayesian}
Yarin Gal and Zoubin Ghahramani.
\newblock Dropout as a bayesian approximation: Representing model uncertainty
  in deep learning.
\newblock In {\em Proceedings of The 33rd International Conference on Machine
  Learning}, volume~48, pages 1050--1059, 2016.

\bibitem{ABDAR2021243}
Moloud Abdar, Farhad Pourpanah, Sadiq Hussain, Dana Rezazadegan, Li~Liu,
  Mohammad Ghavamzadeh, Paul Fieguth, Xiaochun Cao, Abbas Khosravi, U.~Rajendra
  Acharya, Vladimir Makarenkov, and Saeid Nahavandi.
\newblock A review of uncertainty quantification in deep learning: Techniques,
  applications and challenges.
\newblock {\em Information Fusion}, 76:243--297, 2021.

\bibitem{devries2018leveraging}
Terrance DeVries and Graham~W. Taylor.
\newblock Leveraging uncertainty estimates for predicting segmentation quality.
\newblock {\em CoRR}, 2018.

\bibitem{derivation_dropout}
Yarin Gal and Zoubin Ghahramani.
\newblock Bayesian convolutional neural networks with bernoulli approximate
  variational inference.
\newblock In {\em Proceedings of The 6th International Conference on Learning
  Representations}, 2016.

\bibitem{modelformuncertainty_book}
Trevor Hastie, Robert Tibshirani, and Jerome Friedman.
\newblock {\em The elements of statistical learning: data mining, inference,
  and prediction}.
\newblock New York: Springer, 2009.

\bibitem{entropy}
C.~E. Shannon.
\newblock A mathematical theory of communication.
\newblock {\em SIGMOBILE Mob. Comput. Commun. Rev.}, 5(1):3–55, Jan 2001.

\bibitem{vgg}
A.~Dutta and A.~Zisserman.
\newblock The via annotation software for images, audio and video.
\newblock {\em In Proceedings of the 27th ACM international conference on
  multimedia}, 72:2276--2279, 2019.

\bibitem{NEURIPS2020_b5d17ed2}
Le~Zhang, Ryutaro Tanno, Mou-Cheng Xu, Chen Jin, Joseph Jacob, Olga Cicarrelli,
  Frederik Barkhof, and Daniel Alexander.
\newblock Disentangling human error from ground truth in segmentation of
  medical images.
\newblock In {\em Advances in Neural Information Processing Systems},
  volume~33, pages 15750--15762. Curran Associates, Inc., 2020.

\bibitem{NEURIPS2018_473447ac}
Simon Kohl, Bernardino Romera-Paredes, Clemens Meyer, Jeffrey De~Fauw,
  Joseph~R. Ledsam, Klaus Maier-Hein, S.~M.~Ali Eslami, Danilo Jimenez~Rezende,
  and Olaf Ronneberger.
\newblock A probabilistic u-net for segmentation of ambiguous images.
\newblock In {\em Advances in Neural Information Processing Systems},
  volume~31. Curran Associates, Inc., 2018.

\bibitem{mnist}
L.~Deng.
\newblock The mnist database of handwritten digit images for machine learning
  research.
\newblock volume 29(6), pages 141--142. IEEE signal processing magazine, 2012.

\bibitem{mslesion}
O.~Commowick, M.~Kain, R.~Casey, R.~Ameli, J.~C. Ferré, T.~Kerbrat,
  A.and~Tourdias, F.~Cervenansky, S.~Camarasu-Pop, T.~Glatard, S.~Vukusic,
  G.~Edan, C.~Barillot, M.~Dojat, and F.~Cotton.
\newblock Multiple sclerosis lesions segmentation from multiple experts: The
  miccai 2016 challenge dataset.
\newblock {\em Neuroimage}, 244:118589, 2021.

\end{thebibliography}

\end{document}